\documentclass[times,3p]{elsarticle} 

\usepackage[numbers]{natbib}
\usepackage{xcolor}
\definecolor{DarkBlue}{HTML}{005BBB}
\usepackage{booktabs}
\usepackage{multirow}
\usepackage{graphicx} 
\usepackage{amsmath}
\usepackage{lscape} 
\usepackage{tabularx}
\usepackage{tipa}
\usepackage{rotating}

\PassOptionsToPackage{
  colorlinks=true,
  citecolor=DarkBlue,
  urlcolor=DarkBlue,
  linkcolor=black,
  hyperfootnotes=false,
  pdfborder={0 0 0}
}{hyperref}

\usepackage{hyperref} 

\AtBeginDocument{%
  \hypersetup{
    colorlinks=true,
    citecolor=DarkBlue,
    urlcolor=DarkBlue,
    linkcolor=black
  }%
}

\begin{document}

\begin{frontmatter}
\title {A stylometric analysis of speaker attribution from speech transcripts}

\author[1]{Cristina Aggazzotti}
\ead{caggazz1@jhu.edu}

\author[2]{Elizabeth Allyn Smith}
\ead{smith.eallyn@uqam.ca}

\affiliation[1]{organization={Johns Hopkins University},
                city={Baltimore}, 
                state={MD},
                country={USA}}

\affiliation[2]{organization={Universit\'e du Qu\'ebec \`a Montr\'eal},
                city={Montr\'eal},
                state={Qu\'ebec},
                country={Canada}}

\begin{abstract}
Forensic scientists often need to identify an unknown speaker or writer in cases such as ransom calls, covert recordings, alleged suicide notes, or anonymous online communications, among many others. Speaker recognition in the speech domain usually examines phonetic or acoustic properties of a voice, and these methods can be accurate and robust under certain conditions. However, if a speaker disguises their voice or employs text-to-speech software, vocal properties may no longer be reliable, leaving only their linguistic content available for analysis. Authorship attribution methods traditionally use syntactic, semantic, and related linguistic information to identify writers of written text (authorship attribution). In this paper, we apply a content-based authorship approach to speech that has been transcribed into text, using what a speaker says to attribute speech to individuals (speaker attribution). We introduce a stylometric method, \textsc{StyloSpeaker}, which incorporates character, word, token, sentence, and style features from the stylometric literature on authorship, to assess whether two transcripts were produced by the same speaker. We evaluate this method on two types of transcript formatting: one approximating prescriptive written text with capitalization and punctuation and another normalized style that removes these conventions. The transcripts' conversation topics are also controlled to varying degrees. We find generally higher attribution performance on normalized transcripts, except under the strongest topic control condition, in which overall performance is highest. Finally, we compare this more explainable stylometric model to black-box neural approaches on the same data and investigate which stylistic features most effectively distinguish speakers.
\end{abstract}


\end{frontmatter}

\section{Introduction}\label{sec:intro}

Speaker recognition is the task of identifying who is speaking. Methods for performing this task have primarily analyzed phonetic and acoustic properties of the person’s voice without taking other linguistic information, such as the syntax and semantics of their speech, into consideration \cite{gold2019,watt2020}). This is mainly because speaker recognition systems can be highly accurate under certain conditions, with an equal error rate (EER) as low as around 1\% \cite{desplanques2020,ravanelli2021,zeinali2019}. Even in more challenging forensic settings, the EER can still be comparatively low depending on the conditions, such as training using both case-specific and non-case-specific data \cite{morrison2019} and having at least a few seconds of audio \cite{sztah2023}. 

While acoustic methods work well when the audio provides genuine information about the speaker’s voice, we do not have sufficient population statistics to know the extent to which human voices truly differ, nor adequate ways of addressing intra-speaker variation, such as the fact that a person's voice differs at different times of day or under other diverse conditions \cite{garrett1987}. Furthermore, there are several occasions in which the voice might not be available or might be unreliable. For instance, sometimes the audio is not preserved and only a transcript of the speech exists, either due to storage constraints or for easier later examination, such as for virtual meetings. Transcripts are also becoming more prevalent in our daily lives, with both human-generated and, more often, AI-generated transcripts of an ever-increasing amount of spoken news, media, and social communication \cite{cnti2025}. We even see AI-transcription systems advertised for specific domains, such as \emph{Descript} for podcasters or \emph{Otter} for meetings. Finally, if someone disguises their voice, which can now be done easily using a smartphone as most have built-in text-to-speech (TTS) technology, or more complexly with voice cloning or conversion tools (e.g., Google’s StreamVC \cite{yang2024}), voice features are no longer a reliable indicator of who is speaking. All of these factors taken together show there is a greater need in general to understand such a domain of transcribed speech and to be able to accurately identify speakers from speech transcripts. 

While the intention behind using TTS and voice conversion technology is generally practical (e.g., helping those with various visual, verbal, and reading challenges as well as those looking for efficiency or hands-free interactions), bad actors can use the technology maliciously to hide their identity (e.g., in ransom calls), to assume someone else’s identity (e.g., to access a bank account, to obtain personal information), or to distort or change someone’s words (e.g., deepfakes). All of these examples illustrate cases in which voice cannot be used or relied on for speaker identification, and thus current speaker recognition methods will not work. 

Instead, since the only available linguistic information is the words themselves, a different approach that analyzes the content of what is said is needed. Fortunately, authorship attribution models do just that: they examine lexical, syntactic, semantic, and other features of writing to determine who the author is. In forensics, a main method for authorship attribution is stylometry, which measures the frequency of various character, word, token, sentence, and style features in the text, which we describe in detail in~\autoref{sec:model}.\footnote{Here, as in most work in the domain, the following definitions are important. \textit{Characters} are individual Unicode characters, which include letters, digits, punctuation, symbols, and often white spaces (whereas counting letters would limit us to alphabetical characters only). In our work, we consider \textit{words} to only refer to lexical units, whereas \textit{tokens} can be words, punctuation marks, digits, etc. When counting tokens, every instance of that unit is counted, including repetitions; \textit{types}, in contrast, only include distinct units in the text. For example, in the sentence, ``That cat scratched that cat.'', there are six tokens (``that'', ``cat'', ``scratched'', ``that'', ``cat'', ``.''), but only four types, as both ``that'' and ``cat'' appear twice. Note that text is generally lowercased before counting tokens and types.} Despite over a century of research on these stylometric features \cite{lutos1897,neal2017,stamatatos2009}, no tried-and-true set of features has emerged as the most useful across datasets and domains \cite{juola2006,nini2023}, though n-grams and/or the most frequent words have often proven successful \cite{mosteller1963,stamatatos2013}.\footnote{Throughout this article, we adopt the standard notion of an \textit{n-gram} from the natural language processing literature, where it is used to denote a sequence of $n$ contiguous units in a text or in speech (e.g., characters, tokens, etc.). For instance, in the sentence ``I am here'', the token bigrams ($n=2$) are ``I am'' and ``am here''.} Recently, more opaque authorship methods using neural networks have arisen \cite{najafi2022,patel2025,rivera-soto2021,wegmann2022}. Generally, these methods create vector representations of text, called  \textit{embeddings}, that are thought to encode some of these semantic and stylistic features. Although these methods can be quite powerful, analyzing large quantities of data and adapting to new data domains, their decision-making process is not transparent and can be biased in sometimes hard-to-recognize ways, such as due to inherent biases in their training data \cite{solanke2022Explainable}. Judges, attorneys, forensic lab managers, and other forensic scientists similarly show a preference for transparent models \cite{swoffordChampod2022Probabilistic}. Therefore, especially for high-stakes applications within the forensic sciences, a more explainable stylometric model is crucial for both understanding the model’s performance by pinpointing the factors that contribute to its authorship decision and for analyzing large quantities of data that would be untenable manually for a forensic linguist.\footnote{For the purposes of this article, we include any rule-based computational system for authorship in the stylometric category, including those that are largely automated but incorporate more hand-coded linguistic labels or checks, such as Chaski \cite{chaski2005}.}  

It is not a guarantee that authorship models based on textual features would directly capture speech features, though, since speech clearly differs from text. For instance, text often contains punctuation and capitalization, and sometimes misspellings, emoticons, and specialized text formatting, while speech contains restarts (e.g.,~\emph{I ju- I just wanted}) and generally different frequencies and placements of filler words (e.g.,~\emph{umm}, \emph{like}), backchannels (e.g.,~\emph{uh-huh}, \emph{right}), and discourse markers (e.g.,~\emph{well}, \emph{you know}) \cite{duncan1974,sacks1992}. However, previous work has shown that many text-based authorship models do indeed work on (transcripts of) speech \cite{aggazzotti2024,sanjesh2023,sergidou2023,tripto2023}. Although these works tested a range of authorship approaches, and a couple tested some popular stylometric features, they did not use a comprehensive stylometric model or examine which features are most relevant for making attribution decisions. Also, most did not explore how attribution performance is impacted by conversation topic (what its participants are talking about at any given point), which is known to be a confounding factor for authorship \cite{aggazzotti2024,baayen2002,stamatatos2018,wegmann2022}. 

In this paper, we address the following research questions. First, how well do stylometric models distinguish speakers in speech transcripts (\autoref{sec:perf})? Next, how do these stylometric models, which were developed for and tested on text, perform when certain textual cues are lacking, in other words, across different transcription styles (\autoref{sec:style})? Then, to what extent is the stylometric model impacted by conversation topic control (\autoref{sec:topic})? Based on these results, how does the stylometric model’s performance compare to other statistical and neural models on the same data (\autoref{sec:compare})? Finally, which of the stylometric features are most important for distinguishing speakers in this dataset (\autoref{sec:impt})? In addition to providing answers to these research questions, we also furnish the first open-access, explainable stylometric model specifically tailored to attributing speakers from speech transcripts, which can be found here: \url{https://github.com/caggazzotti/styloSpeaker}.

\section{Related Work}\label{sec:related}

The National Institute of Standards and Technology (NIST) has been conducting a Speaker Recognition Evaluation (SRE) since 1996 to encourage research on text-independent speaker recognition \cite{nist}.%
\footnote{In the following discussion of relevant literature, we focus on works that combine authorship and audio information in some way, but there is naturally a much wider range of authorship work on systems applied exclusively to text.} %
NIST extended the 2001 SRE \cite{przybocki2001}, though, to include transcripts of the Switchboard speech corpus, which encouraged the examination of longer-term speech patterns, such as word use, rather than exclusively short-term (i.e., single frame) acoustic features. Doddington \cite{doddington2001} started off this exploration by analyzing unigrams (single units of language, such as a character or word) and bigrams (two consecutive units of language) in the Switchboard speech transcripts, finding that high frequency bigrams performed surprisingly well at detecting speakers. Subsequent work considered other features, such as word-conditioned phone n-grams%
\footnote{Word-conditioned phone n-grams are sequences of consecutive phones found within a particular word, such as the bigrams /k~\textipa{\ae}/ and /\textipa{\ae} n/ found in the word \emph{can}. A system using this feature might count the number of times \emph{can} is pronounced using /k \textipa{\ae}/ and /\textipa{\ae} n/ versus using different phone n-grams like /k \textepsilon/ and /\textepsilon{} n/.} %
\cite{lei2007} and duration-conditioned word n-grams%
\footnote{Duration-conditioned word n-grams involve measuring the time it takes for a speaker to say consecutive sequences of words, such as the duration of saying \emph{I can}, which reveals information about a speaker’s speaking rate and prosody.} %
\cite{tur2007}, as well as combining lexical and acoustic features \cite{campbell2003,kajarekar2003} in an attempt to improve automatic speaker recognition performance. These explorations subsided with the advent of deep learning except in forensic use cases, which often combined acoustic and auditory features with some lexical features, such as idiosyncratic word use \cite{foulkes2012,shriberg2008}.

Analyzing transcripts of speech as a separate but complementary analysis to analyzing speech resurfaced with frequent-word analysis \cite{scheijen2020,sergidou2023,sergidou2024}, which looks at speakers’ use of the most common words, as in some authorship analysis methods. In particular, Sergidou et al. \cite{sergidou2024} combined acoustic features and frequent-word features, finding that fusing the two is especially beneficial when the audio has background noise. In contrast, we focus specifically on how much information can be conveyed through speech transcripts without access to audio. Also, although word frequency is a key stylometric feature, we propose a more extensive stylometry system that not only includes most common words, but also many other features across multiple linguistic levels.  

Other recent studies of speech transcripts include the PAN shared tasks, which are recurring, open competitions on particular topics in digital text forensics and stylometry. The 2023 shared task on authorship verification incorporated speech transcripts, specifically testing cross-discourse authorship performance across written (essays, emails) and spoken (interviews, speech transcripts) discourse types, but their data is not openly available for research \cite{stamatatos2023}. Of the models submitted, two were stylometric in nature \cite{sanjesh2023,sun2023}, but they did not perform as well as the neural embedding-based models; however, all models’ performance was low overall, revealing the difficulty of verification across written and spoken language. 

Tripto et al.\ \cite{tripto2023} tested n-gram and neural authorship attribution models on transcripts of human speech (as well as machine-generated speech transcripts) and found that both character n-gram-based and transformer-based models could effectively distinguish speakers in human speech transcripts (but performed worse on machine-generated speech transcripts). However, Aggazzotti et al.\ \cite{aggazzotti2024} found that when the conversation topic discussed in the transcript is more strictly controlled for, all models’ performance drops to almost chance. This finding aligns with studies using other less precise topic control strategies, such as subreddits of Reddit as proxies for topic, which found that authorship performance decreases with increases in topic control \cite{wegmann2022}. We discuss the effect of conversation topic on attribution performance in more detail in \autoref{sec:data}, but the idea here is that authorship models without some kind of topic control can resort to identifying someone as the person who always talks about, say, cameras in the dataset, whereas using topic control means restricting the data to people talking about cameras, which makes identification harder and forces the model to look for other cues. Due to their focus on topic control, comparison of transcription styles (which is particularly relevant for a stylometric model), and accessible experimental setup on GitHub, we choose to use the Aggazzotti et al.\ \cite{aggazzotti2024} speech transcript attribution benchmark to assess the capabilities of our stylometric model. Note that this kind of replication ability is relatively rare in forensic linguistics as most systems used in investigation and court settings are not transparent, being proprietary or otherwise not made available for testing. In the next section, we discuss the Aggazzotti et al.\ benchmark and our stylometric model in depth.

\section{Method}\label{sec:method}

The Aggazzotti et al.\ \cite{aggazzotti2024} speaker attribution from speech transcripts benchmark uses human-transcribed (rather than automatically-transcribed) transcripts of conversational speech to set a potential ceiling on attribution performance from high-quality transcription. The benchmark tests two baselines and four black-box neural models on the task of speaker verification, in which a trial (i.e., a pair) of speaker transcripts is identified as being from either the same speaker or different speakers. One of the baselines, \textsc{PANgrams} \cite{stamatatos2023}, is in the top two performers, boding well for a more comprehensive stylometric model. This is interesting because \textsc{PANgrams} is a weighted measure of character 4-gram frequency, which means that it uses just one category of feature. The Aggazzotti et al.\ \cite{aggazzotti2024} benchmark specifically focuses on the effect of conversation topic on model performance by creating three difficulty levels based on the amount of conversation topic control applied to verification trials of the speech transcripts. In our experiments, we use the same data and verification setup to provide a clear comparison among all the methods.

\subsection{Data}\label{sec:data}
The speaker verification trials created by Aggazzotti et al.\ \cite{aggazzotti2024} can be obtained using Python scripts available on their GitHub%
\footnote{\url{ www.github.com/caggazzotti/speech-attribution}} %
if you have a subscription to the Linguistic Data Consortium. The trials are of speakers in the Fisher English Training Speech Transcripts corpus \cite{cieri2004}, which contains 11,699 human-transcribed telephone calls between two strangers lasting up to approximately ten minutes each. The corpus is gender-balanced and speakers often participated in multiple calls. 

While it is perhaps a shame that this corpus is not composed of transcripts from calls that have served as evidence in actual judicial proceedings for enhanced forensic validity, we leave such a test for future work for the following reasons, all of which reflect our objectives from the research questions above. First, access to audio data entered into evidence is difficult (often for good ethical reasons), and the data itself is heterogeneous and limited in quantity as well as recording quality. It may not have been transcribed, or, if we worked from transcripts, we may not have access to the audio that would be necessary to verify certain aspects of that transcription; in either case, there would not be multiple transcription styles that were consistent across all samples, which would make testing the effect of transcription style impossible. Second, as will become clear below in this section, the unique setup of the Fisher corpus with assigned conversation topics and repeat callers makes three levels of topic control possible for testing. Without that, it would not be possible to consider the extent to which stylometric models are impacted by conversation topic. Third, the quantity of real forensic data would likely not permit the comparison between stylometric and neural models that is so crucial at this moment as the popularity of the neural approach in artificial intelligence crowds out other types of models, whereas the Fisher corpus provides almost 2000 hours of total audio and corresponding transcripts. Furthermore, using the exact same data as Aggazzotti et al., who provide benchmarks for neural models, allows a direct comparison. 

It is also useful to keep in mind that the innovative nature of this study---testing textual authorship features on transcribed audio---raises the question of what kinds of transcripts we are likely to see in the forensic workflow moving forward. In other words, we know the kinds of texts that have historically been the target of authorship analysis, from suicide notes to threats made online to private text messages, but in the domain of transcribed audio, phone calls, such as 911 calls or the grandparent scam,\footnote{In the grandparent scam, scammers pretend to be a grandchild (or relative) in an emergency situation asking for immediate financial assistance.} are one of the most obvious applications (among others such as audio from bodycams, interviews, etc.). While it is likely that phone calls provided as evidence will have some different features than those in the Fisher corpus, either being between people who know one another, or if between strangers, perhaps representing service encounters, etc., the kinds of features we see in audio overall (restarts, backchannels, pauses, etc.) will be present across call types. In what follows, then, we describe specific features of the Fisher dataset that make it suited to responding to our research questions.

Fisher transcription comes in two different styles: one that resembles written language with features like capitalization and punctuation and another that has been normalized to remove these features. \autoref{tab:transcription} presents a comparison of each style. Crucially, each call was assigned a conversation topic for the speakers to discuss for the duration of the call, which they mostly adhered to \cite{cieri2004}. The topics spanned a variety of areas from hypothetical situations (If you could go back in time and change something, what would it be and why?) to lifestyle choices (Do you prefer eating at a restaurant or at home?) to potentially controversial opinions (Do you think affirmative action is a good policy?).

\begin{table}[ht]
\scalebox{0.85}{\hspace{-.25cm}
\centering
\begin{tabular}{|p{.56\textwidth}|p{.56\textwidth}|}
\hline
\textbf{BBN (text-like)} & \textbf{LDC (normalized)} \\
\hline
L: Hi. [LAUGH] So, do you have pets? & A: hi [laughter] so do you have pets \\
R: Ah, no. & B: (( ah no )) \\
L: Oh. I ha- -- & A: oh \\
R: Do you? & A: i ha- yeah i do i have three dogs [laughter] \\
L: Yeah. I do. I have three dogs [LAUGH] -- & B: (( do you )) \\
R: Oh, okay. & B: oh okay \\
L: -- and I have a bunch of fish. I have -- & A: and i have a bunch of fish i have yeah i have i have a black \\ 
& \hspace{1em}lab he’s eighty pounds big guy and then i have two little dogs \\
& \hspace{1em}like terrier mixes \\
R: Oh. & B: (( oh )) \\
L: Yeah. I have -- I have a black lab; he’s eighty pounds, big guy. And then I have two little dogs, like terrier mixes [LAUGH]. & \\
\hline
\end{tabular}}
\caption{The same transcription sample from each transcription style. BBN (left) has capitalization, punctuation, em-dashes for pauses, and all caps non-speech sounds in brackets. LDC (right) has no capitalization, limited punctuation (hyphens for restarts, apostrophes), non-speech sounds in brackets, and double parentheses around unclear speech hypothesized by the annotator. BBN and LDC sometimes have different segmentations.}
\label{tab:transcription}
\end{table}

Conversation topic is an important variable in attribution work because not controlling for it can make the task too easy and therefore inflate performance scores. For instance, in an authorship verification task in which a model must decide if two documents are written by the same person or not, if documents written by the same author generally discuss the same topic (because a person might tend to talk frequently about a particular topic of interest) and documents written by different authors discuss different topics (because those authors are interested in different things), a model would have an easier time distinguishing authors. Although a scenario like this one is possibly realistic, and conversation topic can be a reliable indicator of authorship in certain instances, that is not always the case, especially in forensic settings. As a simple example, say a homeowner receives a threat note (of unknown authorship) complaining that they do not take care of their yard and the author is suspected to be the nextdoor neighbor. Assuming the neighbor does not have a collection of unsent threat notes, the only available writing samples for comparison to the threat note might be documents discussing entirely different topics (and of different text types and genres), such as work emails or text messages to friends. In this case, a method that “ignores” conversation topic and instead focuses on language or style patterns that pervade a person’s writing regardless of topic (or text type/genre) will be more useful and reliable. 

With that goal in mind, then, we use the assigned conversation topic per call from the corpus to create positive, or same speaker, trials and negative, or different speaker, trials of increasing difficulty. Each trial consists of two halves of a call: all the utterances from a speaker on one call and all of the utterances from a speaker on another call. The ‘base’ level uses no topic control as a baseline, meaning that speakers in the positive and negative trials are matched regardless of whether the topics are the same or different. The ‘hard’ level consists of positive trials in which both sides of the trial involve the same speaker but the call and topic are different, and negative trials in which each side of the trial has a different speaker but the assigned conversation topic is the same. The ‘harder’ level uses the same positive trials as the `hard' setting (same speaker, different topic), but the negative trials have each side of the same conversation as the two transcripts being compared. In other words, these are different speakers on the same call, so they not only discuss the same conversation topic, but also discuss the same subtopics throughout the conversation. One of the reasons we predict that this would make a big difference is because assigned conversation topics can still be pretty broad. For example, if asked to talk about summer plans, one call might focus on students’ part-time jobs, while another might focus on family vacations; although they might share some words in common, such as \textit{July}, there are a number of other words (\textit{shifts}, \textit{flights}, etc.) that could differ substantially, which is less likely to be the case in a single call, where we tend to ask other people follow-up questions about what they say. As an illustration of the various conditions, \autoref{tab:difficulties} provides an example excerpt of negative trials in increasing difficulty in the normalized transcription style.

\begin{table}[t]
\centering
\scalebox{0.89}{
\begin{tabular}{|p{0.5\textwidth}|p{0.5\textwidth}|}
\hline
\multicolumn{2}{|l|}{\rule{0pt}{4ex}\textbf{Different speakers on different topics (easier)}} \\
\hline
well i i never flew or anything before & i don't know i mean i guess uh i \\
(( and i i )) & think thanksgiving might be my favorite holiday \\
surely wouldn't fly now &  how 'bout \\
i'd be afraid to get in a plane or anything & well um  \\
uh & i think because of all of the food and \\
but but you never had a fear of flying & and also thanksgiving is like a very um  \\
 & kind of low key and relaxing holiday for me \\
\hline
\multicolumn{2}{|l|}{\rule{0pt}{4ex}\textbf{Different speakers on the same topic (hard)}} \\
\hline
well i i never flew or anything before  & um do you remember like uh where you were \\
(( and i i )) &  and everything when you heard uh heard about the uh\\
surely wouldn't fly now & attacks   \\
i'd be afraid to get in a plane or anything &  yeah yeah\\
uh & uh-huh  \\
but but you never had a fear of flying & right uh-huh \\
 & oh wow \\
\hline
\multicolumn{2}{|l|}{\rule{0pt}{4ex}\textbf{Different speakers on the same topic in the same conversation (harder)}} \\
\hline
 & no i can't really think of anything that  \\
 &  has really changed at all\\
well i i never flew or anything before &   \\
(( and i i )) & \\
surely wouldn't fly now  & yeah \\
i'd be afraid to get in a plane or anything & right \\
uh & i've flown before but not recently \\
but but you never had a fear of flying &  no not really \\
\hline
\end{tabular}}
\caption{Excerpts of negative (different speakers) trials in increasing difficulty levels (i.e., with more topic control) using the LDC (normalized) transcription style. The topic for all except the top right is, ``What changes, if any, have either of you made in your life since the terrorist attacks of Sept. 11, 2001?''. The topic of the top right is discussing their favorite holiday.}
\label{tab:difficulties}
\end{table}

As seen in \autoref{tab:transcription} and \autoref{tab:difficulties}, there are several notable differences between transcribed speech and written language. For example, depending on the transcription style, there may be non-speech annotations for sounds like laughter, sighing, coughing, and background noise. Also, speakers often start speaking and then rephrase what they were saying (restart), both of which appear in the transcription, whereas in writing, the author would most likely delete the original and only keep the rewritten version. Discourse markers, such as \textit{so} and \textit{oh}, which help manage the flow of the conversation, backchannels, such as \textit{yeah}, \textit{okay}, and \textit{mhm}, which indicate listening or understanding, and filler words, such as \textit{um} and \textit{like}, are all prevalent in speech; they also appear in written language, but their rates of use and sometimes syntactic placement generally differ. There also can be more hedging and repetition in speech than in writing. These differences could make speech transcripts a challenging modality, especially for attribution models that were developed for written data.

As shown in \autoref{tab:stats} below, there are roughly the same number of positive and negative trials for each difficulty level, except in the ‘harder’ setting where the number of negative trials is restricted by the number of conversations. Again, these are the same test set trials from Aggazzotti et al.\ \cite{aggazzotti2024} to enable direct comparison. Each side of a call has approximately 1400 tokens spanning around 100 utterances; however, since the first five utterances often include name and topic introductions, these have been removed (the idea being that the system could rely just on such name/topic information for decision-making rather than other features).

\begin{table}[!h]
\centering
\scalebox{0.9}{
\begin{tabular}{|l|c|c|c|c|}
\hline
 & \textbf{pos trials} & \textbf{neg trials} & \textbf{total trials} & \textbf{speakers} \\
\hline
\textbf{base}   & 956 & 957 & 1913 & 1373 \\
\textbf{hard}   & 959 & 985 & 1944 & 1474 \\
\textbf{harder} & 959 & 558 & 1517 & 1298 \\
\hline
\end{tabular}}
\caption{Number of positive, negative, and total trials as well as the number of speakers per difficulty level for the test dataset.}
\label{tab:stats}
\end{table}

\subsection{Stylometric Attribution Model}\label{sec:model}

Our stylometric speaker attribution model, \textsc{StyloSpeaker}, uses a range of attribution features spanning multiple linguistic levels that come from various sources in the computational linguistics literature (e.g.,~\cite{neal2017,stamatatos2009,strom2021}). Recall that stylometric features were specifically developed for written language so some inherently do not apply to spoken language (or would only reflect the annotator or their style guide), such as the use of digits, the use of American versus British spelling, or the presence of typos and misspellings. Even though punctuation also depends on the annotation style or automatic speech recognition system rather than reflecting the speaker’s style, we keep the punctuation in the text-like transcription style because dashes were often used for hesitation and hyphens for restarts, both of which are characteristic of someone’s speaking style. In this sense, we continue the tradition of trying to find quantifiable proxies for potentially important linguistic features.

For each speaker in a trial, we extracted the range of features shown in \autoref{tab:features}, most of which are stylometric features from previous work. Specifically, we used the following procedure to arrive at this feature set. As a starting point, we extracted the relevant features (for speech transcripts) of successful PAN submissions for the tasks of authorship verification \cite{weerasinghe2020} and style change detection \cite{strom2021,zlatkova2018,zuo2019}, all of which based their features on previous stylometric work, especially the Writeprints feature set \cite{abbasi2008}. These include features related to syntactic parts of speech (POS), in which each token is labeled with its syntactic category (adjective, proper noun, etc.). There are also lexical features, such as the number of contracted and non-contracted words. So-called \textit{weighted} measures are included too, such as Term Frequency–Inverse Document Frequency (TF–IDF), in which there is a comparison between how often a term appears in a document (the Term Frequency) and how rare the term is across the whole text or dataset (the Inverse Document Frequency). 

We also included their readability features measured using Python’s \textsc{textstat}%
\footnote{\url{https://pypi.org/project/textstat/}} %
package. These consist of a number of classic metrics for how easy a text is to read, such as the Flesch Reading Ease \cite{flesch1948}, measuring average sentence length and syllables per word; the SMOG Index \cite{mclaughlin1969}, estimating the years of education required to understand a text based on its number of polysyllabic words; the Gunning Fog Index \cite{gunning1952}, combining average sentence length and the percentage of complex words with three or more syllables; the Coleman–Liau Index \cite{coleman1975}, replacing syllable counts with characters per word and words per sentence to make readability estimation easier for computers; the Linsear Write Formula, created by the U.S. Air Force \cite{linsear1966}, measuring readability for technical manuals by counting easy versus hard words within a 100-word sample; and the Reading Time metric, a heuristic based on 200–250 words per minute, which estimates the approximate time in minutes required to read a given passage; among others.

We made the following changes and additions to the extracted features. First, we replaced NLTK's \cite{nltk} tokenizer and POS tagger with Stanford NLP Group’s Stanza \cite{qi2020} since we found Stanza to perform better (although much more slowly) than both NLTK and spaCy \cite{spacy2020} on the annotation present in the speech transcripts. This reflects a key part of our process, in which we looked at a subset of the data every step of the way, including the results of each kind of tagging. For forensic settings, it is absolutely key to maintain a human-in-the-loop approach for any automated system analyzing forensic data. Second, we augmented Strøm’s \cite{strom2021} 358 function words and 67 function phrases to 390 function words and 69 function phrases by completing paradigms (e.g., only \textit{first}, \textit{second}, and \textit{third} are in the original, but we added up to \textit{tenth}). To measure a speaker’s preference for contracted versus non-contracted words, we augmented Strøm’s \cite{strom2021} list of 29 contractions and their spelled out forms to 61 again by filling in any missing paradigms (e.g., only \textit{that’s} is in the original, but we added \textit{that’ll} and \textit{that’d}). Then, for each speaker in a conversation, we tallied the number of contracted forms and non-contracted forms that were used. 

\begin{table}[]
    \centering
    \scalebox{0.9}{
    \begin{tabular}{|p{2cm}|p{10cm}|}
    \toprule[\heavyrulewidth]
    Character & punctuation mark frequencies (18 total) \\
        & TF-IDF character n-grams (for n = 3, 4, 5, 6)\\ \hline
    Word & average word length (in number of characters) \\
        & ratio of short words ($<$5 chars) to total words (short:W) \\
        & ratio of long words ($\ge$8 chars) to total words (long:W) \\
        & ratio of capitalized words to total words (caps:W) \\\hline
    Token & number of tokens (T) \\
        & number of unique tokens, i.e. types (U)\\
        & ratio of types to tokens (U:T) \\
        & TF-IDF token n-grams (for n = 1, 2, 3)\\ \hline
    Syntax & number of sentences \\
        & average sentence length (in number of tokens) \\
        & function word frequencies (390 words) \\
        & function phrase frequencies (69 phrases)\\ 
        & POS tag frequencies (using Stanza, UPOS tagset)\\ 
        & TF-IDF POS tag n-grams (for n = 1, 2, 3)\\  \hline
    Discourse & vocabulary richness (Yule's \emph{I})\\ 
        & readability measures (9 total; using Python's \textsc{textstat})\\ 
        & ratio of hapax legomena to total number of words \\
        & ratio of hapax dislegomena to total number of words  \\ 
        & number of contracted terms (out of 61 total)\\ 
        & number of non-contracted terms (out of 62 total)\\
    \bottomrule[\heavyrulewidth]
    \end{tabular}}
    \caption{Stylometric features by linguistic level. Words include only lexical units, while tokens include lexical units as well as punctuation and any special characters. Total word count (W) was not included as a feature because the number of tokens (T) conveyed nearly the same information (since the word and token count would be the same if there were no punctuation marks/symbols/digits or only slightly higher if there were).}
    \label{tab:features}
    \end{table}

We additionally included the following static\footnote{\textit{Static features} are fixed and are measured directly from the text, such as function word frequencies and average sentence length. \textit{Dynamic features}, on the other hand, vary depending on the dataset and have potentially very large feature spaces; n-grams, for example, are dynamic because certain n-grams might be in one dataset but not another (e.g., \textit{mput} would appear in a text talking about computers but might not appear in other texts).} %
features: punctuation mark frequencies (18 total), token count, unique token count (i.e., types), average word length (in number of characters), number of sentences, average sentence length (in number of tokens), POS tag frequencies (using the Universal POS tagset), the ratio of hapax legomena and hapax dislegomena to the total number of words \cite{weerasinghe2020}, and Yule’s \textit{I}%
\footnote{\url{https://gist.github.com/magnusnissel/d9521cb78b9ae0b2c7d6}}\textsuperscript{,}%
\footnote{While a full calculation of Yule's I is beyond the scope of this work, suffice it to say that Shakespeare would have a low value, reflecting a broad, less-repeated vocabulary; a police interrogation might have a very high value, with many of the same words repeated; and a local newspaper might be somewhere in between.} %
to measure vocabulary richness. Adapting from Altakrori et al.\ \cite{altakrori2021}, we included the following word ratios: the ratio of the number of short words ($\leq 5$ characters) to the total number of words, the ratio of the number of long words ($\geq 8$ characters) to the total number of words, the ratio of types (unique words) to the total number of words, and the ratio of the number of capitalized words to the total number of words. Although unlikely to make a big difference, to err on the side of being more precise, we specified a distinction between token-based (words, punctuation, symbols) and word-based (lexicon only) features. For example, we only considered actual lexical words when counting word length, following the linguistic notion of what counts as a word, despite the fact that most previous (sometimes non-linguistically-informed) work considered all tokens, even punctuation. Also, when calculating the word and hapax ratios, we use the total number of words, not tokens. For hapax legomena, we count the number of lexical words that only appear once in the speaker’s utterances for a call and for hapax dislegomena, the number that appear twice.  

In addition to all of these static features, which are fixed features across all texts, we used dynamic features, namely character, token, and POS tag n-grams, which vary depending on the text and can thus have extremely large feature spaces. To help reduce the feature space, we selected the top 2000 of each kind of n-gram ($max\_features=2000$) based on their TF-IDF score (using scikit-learn's \textsc{TfidfVectorizer}). (We discuss this and other design decisions next.) Based on previous work, we used $3 \leqslant n \leqslant 6$ for character n-grams, $1 \leqslant n \leqslant 3$ for token n-grams, and $1 \leqslant n \leqslant 3$ for POS tag n-grams. We ignored any n-gram that appeared in less than 10\% of the training transcripts ($min\_df=0.1$) since it was likely to be a transcription error or specific to a particular speaker (but speakers in the training and test sets were distinct).%
\footnote{Large numbers of these rare n-grams might cause the logistic regression classifier (discussed in \autoref{sec:logreg}) to overfit on the training data and thus not generalize well to the test data. The risk of overfitting here is the possibility that by giving the classifier too many features from the training data, it might essentially memorize the data rather than being forced to learn more general patterns, which would make it perform well on the training data that it had memorized, but poorly on the new test data. Future work could explore various settings for the minimum number of transcripts an n-gram has to appear in to be included, including different settings per kind of n-gram.}\textsuperscript{,}%
\footnote{Note that the set of n-grams only comes from the training data, not the test data, because the \textsc{TfidfVectorizer} extracts the relevant n-grams from the training data and then counts how often they occur in the test data; therefore, this way of measuring n-grams does not capture idiosyncratic phrases used by individual speakers, such as the Unabomber's ``you can't have your cake and eat it too.'' \cite{Kaczynski1995Manifesto, Pullum2005Cake}} %
We also used L2 normalization ($norm=$`$l2$'), which rescales the feature vector, to limit the effect of different text/conversation lengths. In other words, one sample could have more tokens of the word \textit{the} than another just because it is twice as long, but the normalization allows us to essentially put them on the same scale to compare apples to apples. To further illustrate how the features work, we walk through an example for the utterance ``Leave \$50{,}000 in the dumpster.'' in \ref{app:feattable}, which also reveals that for some extracted features, \textit{\$50,000} is considered a word, and in others, it is not. 

After obtaining values for each feature, we scaled the non-n-gram features using Python’s scikit-learn’s \textsc{StandardScaler} to transform the individual numerical range of each stylometric feature to a uniform range across features and to ensure the stylometric features were not underweighted in comparison to the TF-IDF features (which are already normalized). The features were extracted from each side of a trial and then combined (e.g., by taking their absolute difference) to produce one feature vector per trial. We discuss this procedure in more detail in \autoref{sec:exp}. 

\paragraph{Experimental design decisions}
We experimented using the top $1000$, $2000$, $3000$, $5000$, and $7000$ TF-IDF n-gram features and found diminishing returns with more features. For computational efficiency and minimizing the risk of overfitting, we chose 2000 features as a balance between performance and computational cost. We also experimented with $2 \leqslant n \leqslant 7$ for character n-grams, $1 \leqslant n \leqslant 5$ for token n-grams, and $1 \leqslant n \leqslant 5$ for POS tag n-grams but did not get as high of performance with the larger values for $n$, most likely because the additional larger n-grams are less common, so they could add noise that the classifier might learn and thus not generalize well to new data (i.e., it might create another possibility of overfitting the training data). 

\subsection{Logistic regression classifier}\label{sec:logreg}
After calculating counts of all of the features mentioned above for each text, which are stored in vector representations, we then trained a binary logistic regression classifier (with a maximum of 1000 iterations) on these feature vectors and assessed its ability to identify each trial (pair of single conversation sides) as being said by the same speaker or different speakers.%
\footnote{Note that this research question differs from the likelihood ratio framework, which would ask how probable a trial is under the hypothesis that the speakers are the same versus the hypothesis that they are different. The likelihood ratio would not be applicable here because we are asking if the speakers can be discriminated from each other and, more importantly, our features are too varied (e.g., n-gram TF-IDF scores, frequency counts, ratios), numerous (thousands of n-grams), and correlated (e.g., token n-grams could overlap with function words/phrases) to fit the distributional assumptions required by the likelihood ratio. For examples of using the likelihood ratio for authorship attribution, see \citet{ishihara-2021-score,ishihara2022likelihood}.} 
An advantage to using logistic regression is the ability to examine the importance of each feature (via their coefficient) in making the speaker verification decision. The classification results are then evaluated using the area under the receiver operating characteristic curve (AUC), a measure of how well a model can distinguish between classes (e.g., positive and negative trials for our case of binary classification). AUC is between $0$ and $1$, with $1$ being perfect discrimination between classes and $0.5$ being chance performance (i.e., random guessing). AUC is a commonly used metric for classification because, unlike accuracy, it is robust to imbalances in the number of classes and considers multiple thresholds between classes (rather than arbitrarily picking one like accuracy does, e.g., below $0.5$ is classified as a negative trial and above $0.5$ is classified as a positive trial). We include the results using two other metrics, accuracy and equal error rate, for comparison in \ref{app:acceer}. 

Overall, the approach presented here renders a more holistic representation of the input text (with slightly more linguistically-informed features in addition to basic stylometric ones), and it has a more interpretable analysis than other black-box authorship models due to its use of features that have been tested in previous literature. Put another way, logistic regression itself is a transparent statistical model (an interpretable classifier), but it can be a component of a transparent authorship system or an opaque authorship system depending upon the nature of its input: with human understandable feature definitions like those used here, we can interpret the coefficients and explain the classifier decisions in linguistic terms. With a pipeline of opaque features, such as many-dimensional embeddings, one loses interpretability and transparency.

\section{Experiments}\label{sec:exp}

We present the results from each experiment corresponding to the order of research questions in~\autoref{sec:intro}. 

\subsection{Stylometric performance on speech transcripts}\label{sec:perf}
In order to assess how well \textsc{StyloSpeaker} performs on the Fisher speech transcript trials, we experimented with four different ways of combining the features within each trial for input to the logistic regression classifier. Verification involves pairs of documents, each document with a long list of feature values, but the classifier requires a single input. Therefore, the features from each side of the trial need to be combined in some way to produce one feature vector. \emph{Concat} concatenates the feature vectors from each side of the trial to form one feature vector for the trial. Although this approach includes feature values for both sides of a trial, it does not take into account the relationship between the features for each side. \emph{Diff} does consider the relationship between the documents in a trial by taking the absolute difference between the features of each side of a trial.%
\footnote{We also tried just taking the difference (not the absolute difference) and results were strictly worse.}
\emph{Diff + Prod} takes both the absolute difference between the feature vectors and the product of the feature vectors for each trial to see if combining the features another way will provide more information that can be used by the classifier. \emph{Concat + Diff} takes both the concatenated feature vectors per trial as well as the absolute difference feature vector for each trial. 

\begin{table}[h!]    
\centering
\scalebox{0.9}{\hspace{-.3cm}
    \begin{tabular}{|c|cccc|cccc|}
    \toprule
  \multirow{2}{*}{AUC $\uparrow$} & \multicolumn{4}{c|}{\bf BBN} & \multicolumn{4}{c|}{\bf LDC} \\
 & \it Concat & \it Diff & \it Diff $+$ Prod & \it Concat $+$ Diff & \it Concat & \it Diff & \it Diff $+$ Prod & \it Concat $+$ Diff \\ \toprule
\bf Base & 0.537 & \underline{0.858} & 0.818 & 0.773 & 0.535 & \bf 0.861 & 0.824 & 0.777\\
\bf Hard & 0.550 & \underline{0.826} & 0.815 & 0.716 & 0.556 & \bf 0.829 & 0.820 & 0.741\\
\bf Harder & 0.513 & \bf 0.893 & \underline{0.887} & 0.807 & 0.518 & 0.862 & 0.860 & 0.781\\\bottomrule
    \end{tabular}}
    \caption{AUC performance across four feature measurements on all levels of topic control for text-like BBN (left) and normalized LDC (right) transcription trials. Best performance per difficulty level (across both encodings) is bolded, second-best underlined.}
    \label{tab:measure}
\end{table}

The first row of the left side of \autoref{tab:measure} shows the AUC results across all four feature measurements for the text-like BBN transcription in the ‘base’ setting (no topic control). Taking the absolute difference of feature vectors (\emph{Diff}) produces the highest AUC score. Concatenating the feature vectors produces the lowest scores, most likely because the individual feature values do not provide information about how the texts relate to each other that would be helpful in determining whether they were said by the same person or not. On the other hand, the combined feature measurements, \emph{Diff + Prod} and \emph{Concat + Diff}, might provide too much information and the classifier might start overfitting the data. As a reminder, overfitting happens when the classifier learns the training data too closely, including all of its specificities, and then does not generalize well to new data that might have its own particularities. Nonetheless, using \emph{Diff}, the stylometric method is able to distinguish speakers fairly well.  

\subsection{Transcription style impact}\label{sec:style}
Turning to the first row of the right side of \autoref{tab:measure} are the results across all four feature measurements for the normalized LDC transcription (no capitalization and limited punctuation) in the ‘base’ setting. Again, results on the absolute difference of the feature vectors are best. Comparing the results of the two encodings, the text-like BBN transcription performs slightly worse than the normalized LDC transcription, which is perhaps surprising considering the stylometric features were developed for written language. However, only a few of the features selected actually depend on textual features like capitalization and punctuation, and the majority of the total features overall are n-grams, which are lowercased. Therefore, transcription style does play a role, albeit somewhat minor, in stylometric attribution performance on this dataset. 

\subsection{Topic manipulation impact}\label{sec:topic}
The remaining two rows in \autoref{tab:measure} show the AUC scores for the four feature measurements of both transcriptions for the harder levels of topic control. The absolute difference in features (\emph{Diff}) remains the overall best performer for both encodings. The LDC transcription continues to perform marginally better than the BBN transcription in the ‘hard’ setting, but BBN shows a much bigger improvement over LDC in the ‘harder’ setting. Overall performance is also highest on the ‘harder’ setting, suggesting that especially in difficult topic manipulation conditions, the stylometric features succeed in providing distinguishing information about the speakers. The results on each difficulty level using the metrics of accuracy and equal error rate for \emph{Diff} are in \autoref{tab:acceer} in~\ref{app:acceer}.

\subsection{Comparison to other models}\label{sec:compare}

\begin{table}[h!]
\centering
\scalebox{1}{
\begin{tabular}{c c}
\rotatebox[origin=c]{90}{\textbf{BBN}} \hspace{-.2cm}
&
\begin{tabular}{|c|ccc|ccc|}
\toprule
\multirow{2}{*}{AUC $\uparrow$} 
  & \multicolumn{3}{c|}{\textit{Explainable models}}
  & \multicolumn{3}{c|}{\textit{Neural models}} \\
& \bf Stylo & \bf LFTK & \bf PANgrams & 
  \bf SBERT & \bf CISR & \bf LUAR \\ \midrule
\bf Base   & \bf 0.858 & 0.665 & 0.755 & 0.689 & 0.663 & \underline{0.764} \\
\bf Hard   & \bf 0.826 & 0.679 & 0.633 & \underline{0.809} & 0.619 & 0.801 \\
\bf Harder & 0.893 & 0.833 & 0.419 & \bf 0.936 & 0.864 & \underline{0.909} \\
\bottomrule
\end{tabular}
\end{tabular}}

\scalebox{1}{
\begin{tabular}{c c}
\rotatebox[origin=c]{90}{\textbf{LDC}}  \hspace{-.2cm}
&
\begin{tabular}{|c|ccc|ccc|}
\toprule
\multirow{2}{*}{AUC $\uparrow$} 
  & \multicolumn{3}{c|}{\textit{Explainable models}}
  & \multicolumn{3}{c|}{\textit{Neural models}} \\
& \bf Stylo & \bf LFTK & \bf PANgrams & 
  \bf SBERT & \bf CISR & \bf LUAR \\ \midrule
\bf Base   & \bf 0.861 & 0.679 & 0.762 & 0.694 & 0.722 & \underline{0.844} \\
\bf Hard   & \bf 0.829 & 0.678 & 0.623 & \underline{0.830} & 0.641 & 0.872 \\
\bf Harder & 0.862 & 0.787 & 0.416 & \bf 0.935 & 0.781 & \underline{0.894} \\
\bottomrule
\end{tabular}
\end{tabular}}
\caption{AUC performance across explainable and neural models on all levels of topic control for BBN (top) and LDC (bottom). Best performance is bolded; second-best underlined. All differences among models per difficulty level within each encoding are statistically significant ($p < 0.0001$).}
\label{tab:all}
\end{table}

We include a mix of explainable and neural models for comparison with the \textsc{StyloSpeaker} results. We choose two explainable models. \textsc{LFTK}\footnote{\url{https://github.com/brucewlee/lftk}} is a feature extraction toolkit that only uses static features, many of which are also included in \textsc{StyloSpeaker}, but does not include n-grams \cite{lee2023}.\footnote{We included all possible features, but future work could explore different subsets of features.}  Once the features are extracted, we use the same pipeline as for \textsc{StyloSpeaker}, taking the absolute difference between features in trials and fitting a logistic regression classifier on those features to obtain predictions of whether the speaker is the same or not. As a counterpart to this feature-only model, we also use an n-gram only model, namely \textsc{PANgrams} from Aggazzotti et al.\ \cite{aggazzotti2024}. \textsc{PANgrams} is the PAN competition authorship verification baseline and uses TF-IDF-weighted character 4-grams \cite{stamatatos2023}. 

For the neural models, we select only the models that Aggazzotti et al.\ \cite{aggazzotti2024} found to perform best: Sentence-BERT (SBERT),\footnote{\url{huggingface.co/sentence-transformers/all-MiniLM-L12-v2}} Content-Independent Style Representations (CISR),\footnote{\url{huggingface.co/AnnaWegmann/Style-Embedding}} and Learning Universal Authorship Representations (LUAR).\footnote{\url{huggingface.co/rrivera1849/LUAR-MUD}} SBERT focuses on lexical co-occurance as a proxy for semantics, creating semantically-related sentence embeddings of text. In contrast, CISR focuses on author style and intentionally aims to ignore semantic relationships in an attempt to be content-agnostic. Compromising between these two, LUAR aims to capture author style and some semantics without being too sensitive to content. Although these neural models are powerful and adaptable, they are black boxes and thus it is not clear exactly what information is used in their decision-making. As a result, their decisions could be biased by the data the model was trained on or they could perform better on some datasets than others, potentially making their results misleading and inaccurate. 

For easier comparison, we choose one \textsc{StyloSpeaker} model based on the highest performing feature measurement, \emph{Diff}. We then compare this model’s AUC performance with the fine-tuned versions of the other models. Note that the models from the previous work involved fitting a multilayer perceptron classifier on the training trials whereas the stylometric model involved fitting a logistic regression classifier. Recall that we used a logistic regression classifier since the model’s coefficients are easily interpretable and provide information of how important each feature is for the attribution decision (see \autoref{sec:impt}).

The AUC results across all models for the three topic control settings for the BBN (top) and LDC (bottom) transcriptions are in \autoref{tab:all}. Based on a paired t-test, all differences among models per difficulty level within each encoding are statistically different ($p < 0.0001$). On the ‘base’ and ‘hard’ difficulty levels, \textsc{StyloSpeaker} performs the best for both the BBN and LDC transcriptions, while in the ‘harder’ setting, the neural model SBERT has an advantage. Although the ‘harder’ setting is useful for creating a difficult topic manipulation setting, it is less common in real-world cases. Thus the fact that the stylometric system, which is less prone to topic manipulations, performs best in the ‘base’ and ‘hard’ settings is promising for explainability and using these systems in the courtroom. 

\textsc{StyloSpeaker} significantly outperforms the other two explainable models, which rely on either static features or (character) n-grams but not both. To further assess the role of static and dynamic features, we next analyze which features are most useful for making speaker verification decisions at each difficulty level.

\subsection{Most important features}\label{sec:impt}
Looking at the coefficients of logistic regression reveals which features were most important for the classifier to decide if two sides of a trial were spoken by the same speaker or different speakers, with larger absolute coefficients indicating higher importance. Taking the absolute value of the coefficients yields the features with the most predictive power overall, while considering the sign of the coefficient reveals whether features are predictive of same speaker trials (positive coefficient) or different speaker trials (negative coefficient). 

The following figures (\ref{fig:base}-\ref{fig:harder}) show two plots for each difficulty level. Once again focusing on \emph{Diff}, the first is a heatmap ranking the 20 most important absolute features for the BBN transcription (left) and LDC transcription (right). The heatmap color scale ranking ranges from dark blue, indicating the most important feature, to light yellow, indicating a less important feature. The colors are sorted from most to least significant for BBN (left) for ease of viewing and comparison with LDC (right). Any features that are of top significance in one but not in the other are white (no color fill). The feature names are composed of two parts separated by an underscore: the first part is whether the feature is a character n-gram (“char”), token n-gram (“tok”), POS tag n-gram (“pos”), or static stylometric feature (“stylo”), and the second part is the feature itself (for n-grams) or the feature name. The second plot shows the top 15 signed (i.e., not absolute) top features that are discriminative of negative (different speaker) trials, and the top 15 that are discriminative of positive (same speaker) trials. They are in decreasing order based on absolute importance. Therefore, the top 15 overall features on this plot align with the top 15 in the heatmap, but the remaining 15 may or may not align depending on whether they have the highest absolute coefficient or not.

The most common feature category overall across difficulty levels and transcriptions is token n-grams, and generally token unigrams. With the exception of in the ‘base’ level, several of these tokens are surprisingly not function words (e.g., \emph{friends}, \emph{family}, \emph{computer}). These results differ from previous work that has found function words and character n-grams to be particularly effective at distinguishing authors \cite{grieve2007,houvardas2006,keselj2003,kestemont2014,mosteller1963}; however, Sari et al.\ \cite{sari2018} found that character n-grams only perform well on certain kinds of datasets and word n-grams are best for datasets with more topical diversity, such as our ‘hard’ and ‘harder’ datasets, aligning with our results. 

\begin{figure}[!ht] 
       \centering
      \includegraphics[width=.8\linewidth]{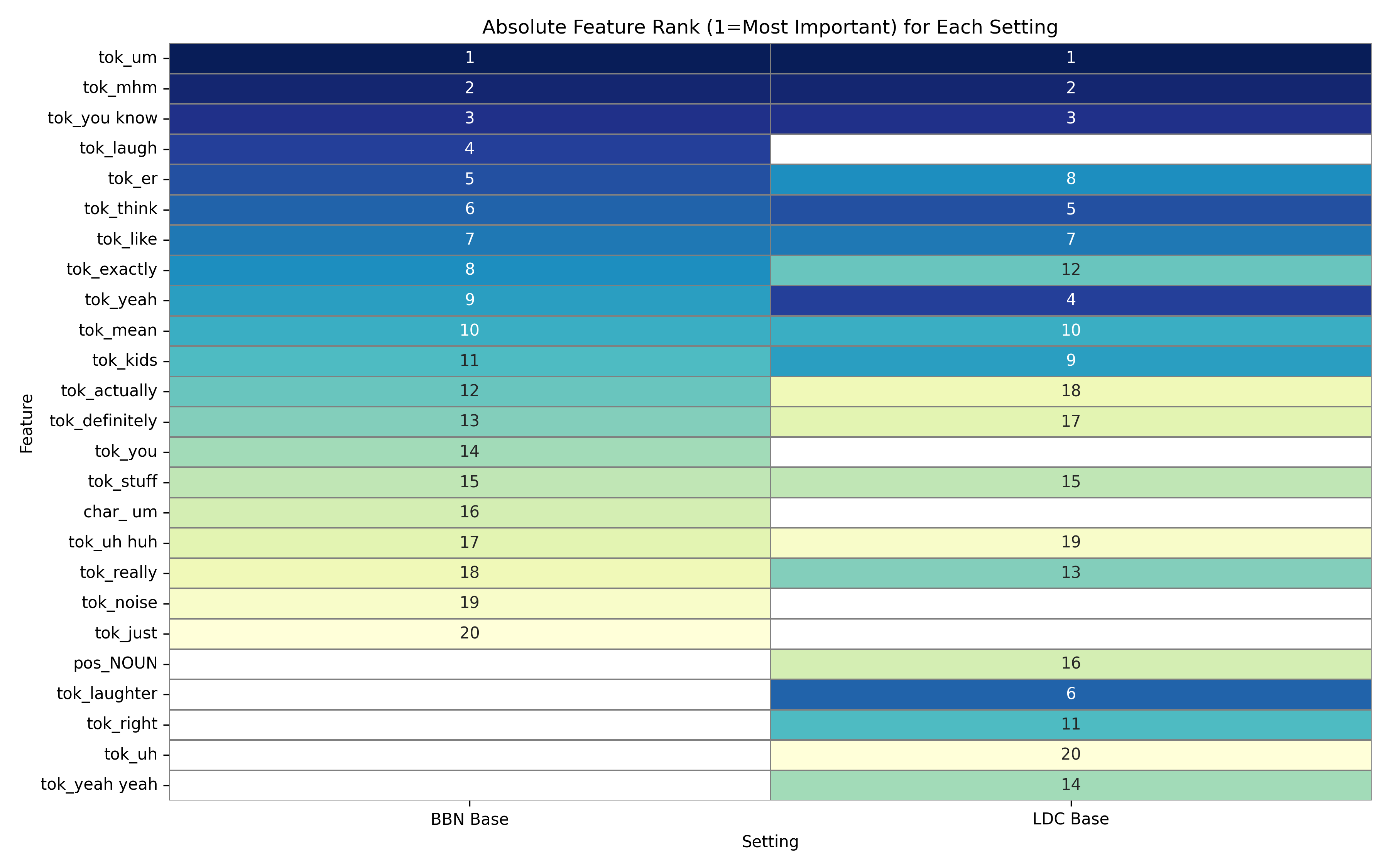}\\ \vspace{-.16cm}
      \includegraphics[width=0.49\linewidth]{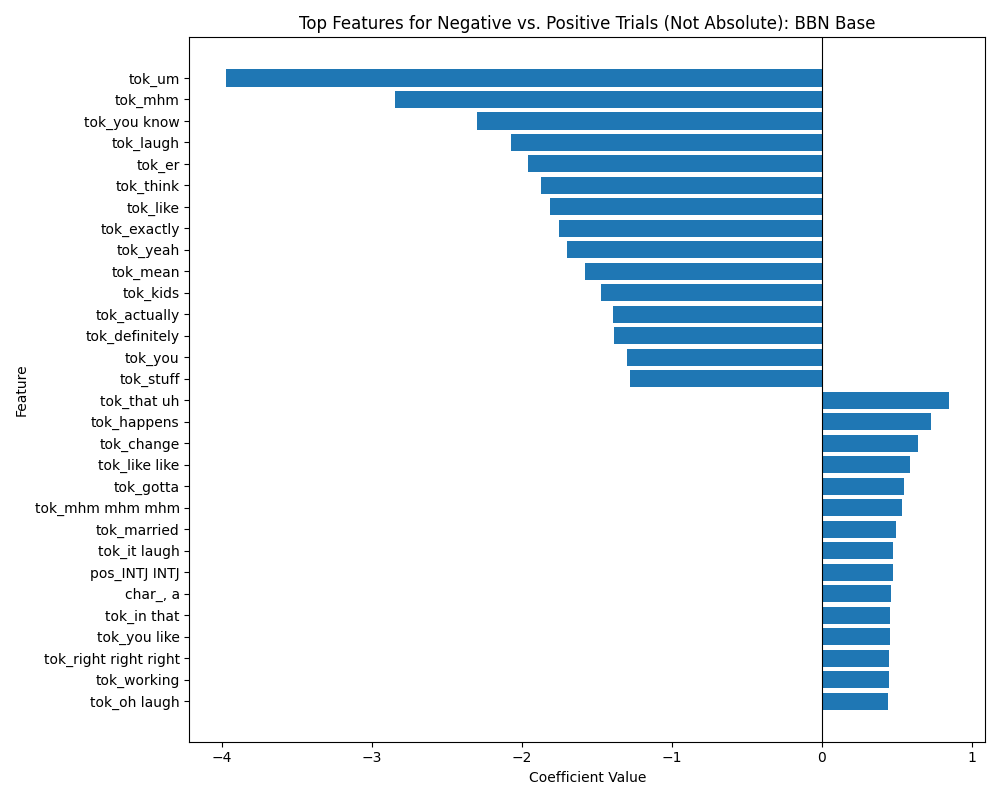}
      \includegraphics[width=0.49\linewidth]{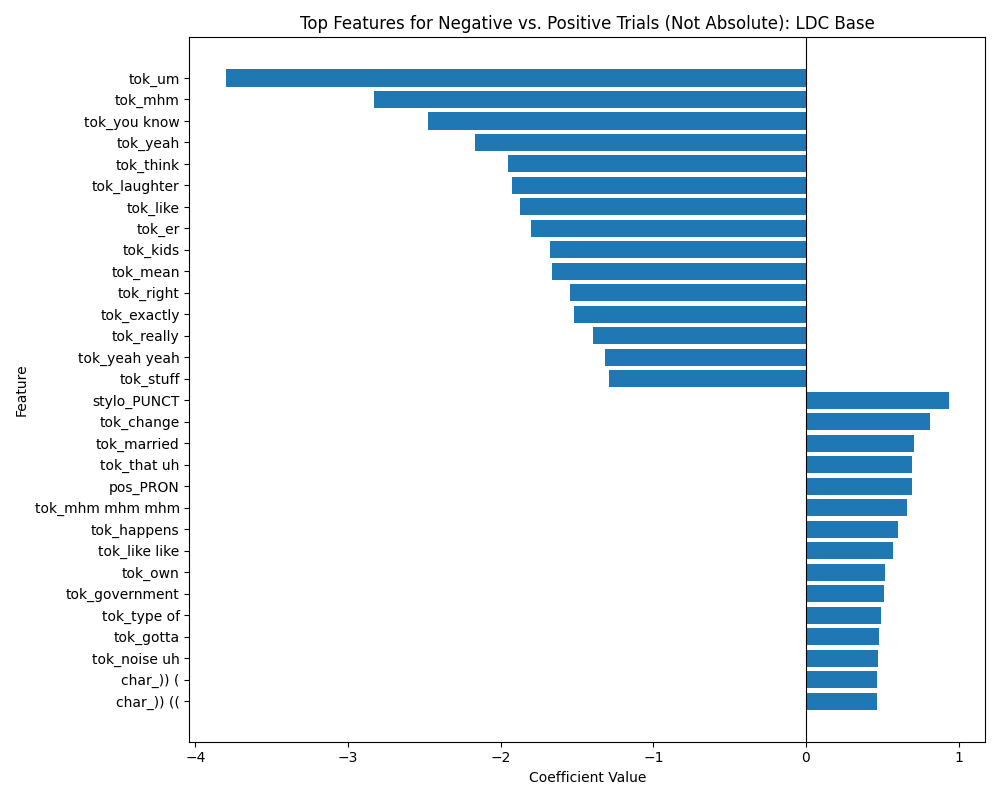}\vspace{-.2cm}
       \caption{\textbf{\emph{Top}}: Heatmap for the BBN (left) and LDC (right) transcription in the ‘\textbf{base}’ difficulty level showing the ranking of the 20 most important absolute (not based on the logistic regression coefficient’s sign) features for the speaker verification task. For both transcriptions, token unigrams, especially those characteristic of speech, are most important. \textbf{\emph{Bottom}}: Absolute ranking of the top 15 features for distinguishing negative (coefficients with a negative sign) and positive (coefficients with a positive sign) trials for BBN (left) and LDC (right). Overall, features for distinguishing negative trials have larger absolute coefficients and are thus more predictive. 
        }
        \label{fig:base}\vspace{-.3cm}
   \end{figure}

Looking in more detail at each difficulty level, in the ‘base’ level heatmap in \autoref{fig:base}, many of the top token features for both transcriptions involve function words, unlike in the ‘hard’ and ‘harder’ settings, and are characteristic of speech, such as filler words (\emph{um}, \emph{er}, \emph{like}), backchannels (\emph{mhm}, \emph{exactly}, \emph{yeah}), discourse markers (\emph{you know}), and non-speech sounds (\emph{laugh} in BBN, \emph{laughter} in LDC based on transcription style). Differences between the two transcription styles include that the token n-gram ``you'' is relevant for BBN but not for LDC, and the POS tag n-gram, NOUN, is relevant for LDC but not for BBN. 

The next two plots in this figure show the top 15 features that are discriminating of negative trials and positive trials for BBN (left) and LDC (right). Overall, the coefficients for the negative trials are larger than those for the positive trials, indicating that features predictive of different speakers had greater weight than those predictive of the same speaker for the logistic regression classifier. In fact, many of these features are likely to differ by speaker, such as whether a speaker uses the filler word \emph{um} or \emph{er} or laughs a lot, and thus it makes sense that these would help distinguish speakers. Of the negative predictive features, BBN and LDC share most of them albeit in slightly different orders, except BBN has \emph{actually}, \emph{definitely}, and \emph{you}, while LDC has \emph{really}, \emph{right}, and \emph{yeah yeah}. It is difficult to pinpoint an exact reason for these differences, but it could be due to different transcriptions or tokenizations. For the positive predictive features, there are more differences between the transcription styles. Also note that there is a gap between the absolute values of the last negative and first positive feature because there are several other intervening negative features in the overall ranking of top features. (This is not the case in ‘hard’, for instance, as we will see next.) BBN has an additional backchannel (\emph{right right right}) and also INTJ INTJ, which could be repeated backchannels or filler words, while LDC has more punctuation (PUNCT, double parentheses). Recall that the LDC transcription is normalized to remove all punctuation other than hyphens, apostrophes, and double parentheses, which denote unclear speech that the annotator hypothesized to be correct. The double parentheses thus could be revealing of a speaker if they mumble or there is background noise impacting the quality of the audio. 

\begin{figure}[!ht]
       \centering
       \includegraphics[width=.8\linewidth]{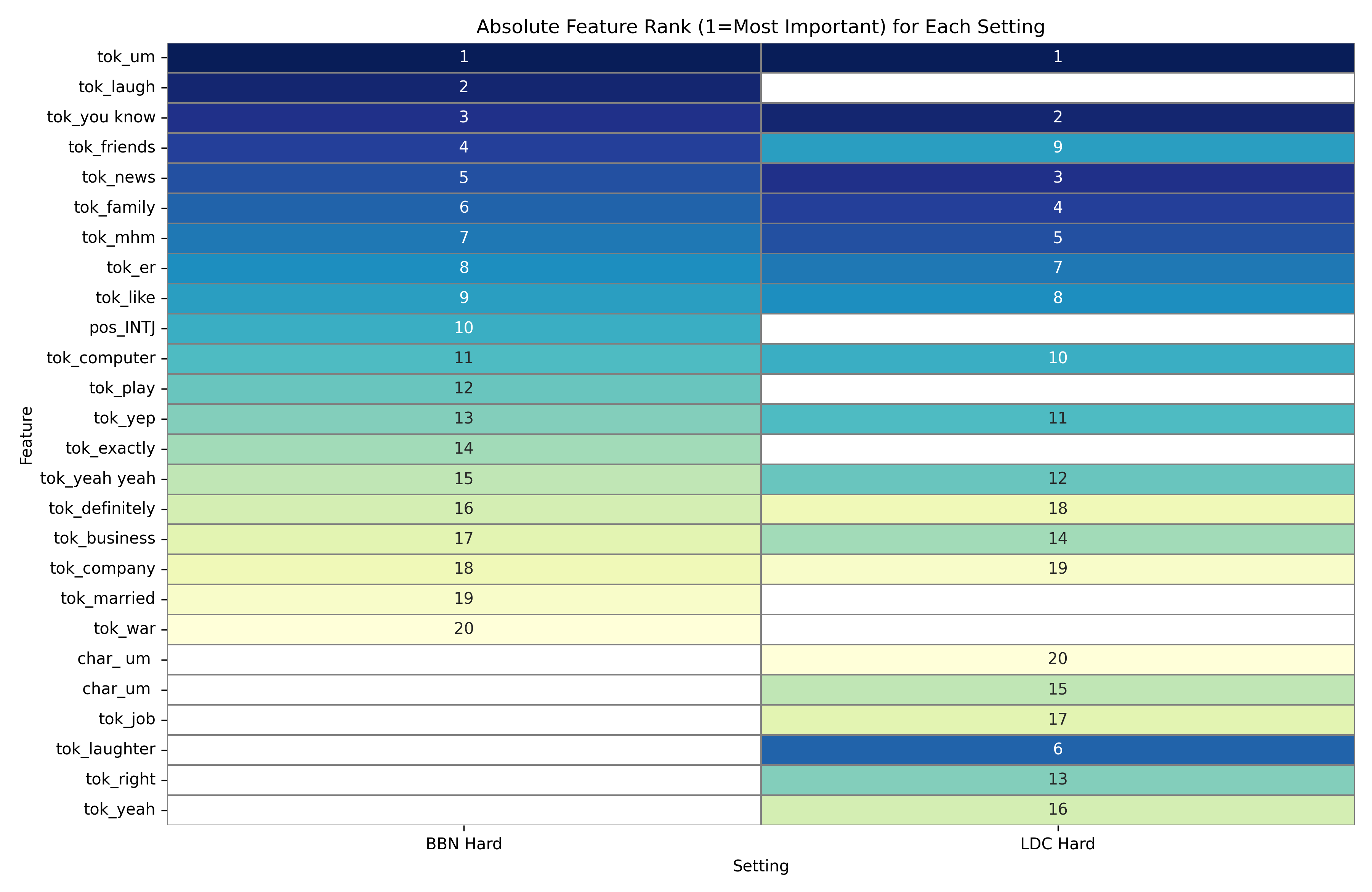}\\ \vspace{-.16cm}
      \includegraphics[width=0.49\linewidth]{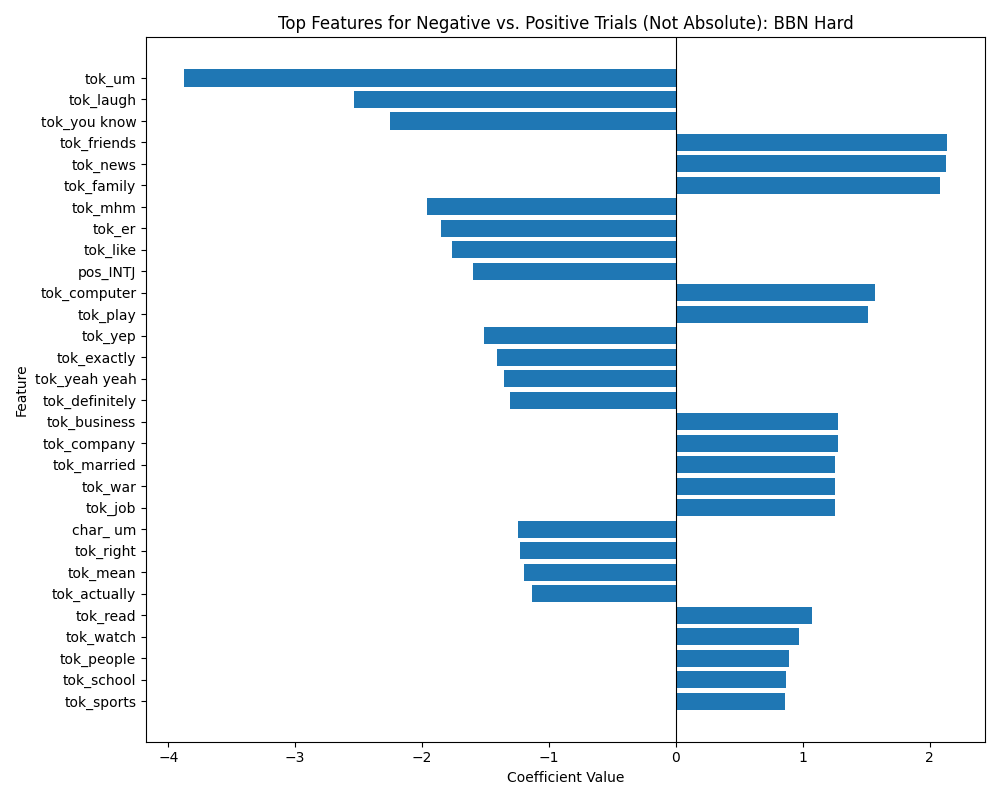}
      \includegraphics[width=0.49\linewidth]{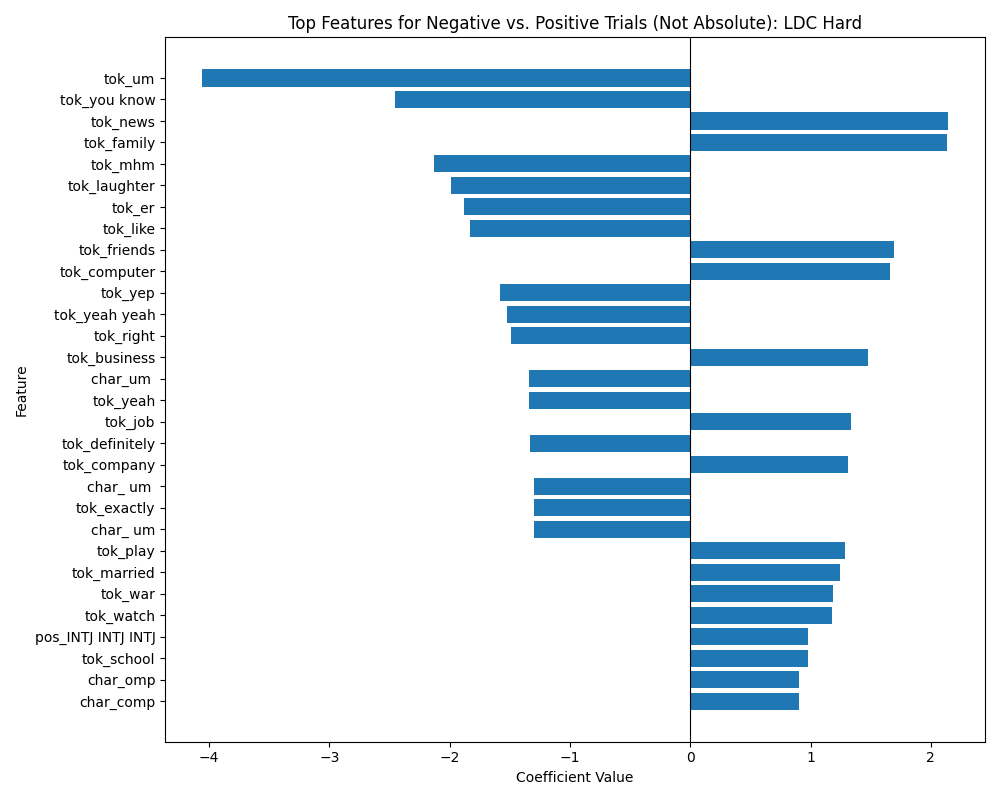}\vspace{-.2cm}
       \caption{\textbf{\emph{Top}}: Heatmap for the BBN (left) and LDC (right) transcription in the ‘\textbf{hard}’ difficulty level showing the ranking of the 20 most important absolute (not based on the logistic regression coefficient’s sign) features for the speaker verification task. For both transcriptions, token unigrams, including both function and “content” words, are most important. \textbf{\emph{Bottom}}: Absolute ranking of the top 15 features for distinguishing negative (coefficients with a negative sign) and positive (coefficients with a positive sign) trials for BBN (left) and LDC (right). Overall, both negative and positive features tend to have large absolute coefficients and thus are both predictive.  
        }
        \label{fig:hard}\vspace{-.3cm}
   \end{figure}

In the ‘hard’ level in \autoref{fig:hard}, in addition to the same function words that were most important in the ‘base’ level, there are now also several content words (e.g., nouns, main verbs), such as \emph{friends}, \emph{news}, and \emph{family}, for both BBN and LDC transcriptions. Since this setting has some topic control, it may seem surprising that content words are more discriminating of speaker. However, looking at the breakdown plots (bottom of \autoref{fig:hard}) for which features are more predictive of negative versus positive trials, we see that all of these content words are used to discriminate the same speaker trials in which the speakers are discussing different topics. Since we are taking the absolute difference of the features, if the speaker is discussing different topics, the content words will differ, creating a larger difference in those features. During training, logistic regression ``learns'' that large differences in content features correlate with same speaker trials and thus content features are given a large positive coefficient (or importance weight). For different speaker trials, the speakers are discussing the same topic, so ostensibly have more content word overlap and thus a smaller difference in those features; instead, function word or style features might have larger differences. Therefore, the logistic regression learns to associate large differences in function features with different speaker trials and those features are given a large negative weight. As a result, the logistic regression is likely using conversation topic as a clue for classifying the trials when those features are useful. However, performance on the ‘hard’ level is not as high as on the ‘base’ level ($0.829 < 0.861$ on LDC) probably because the topic manipulation is imperfect (e.g., a speaker might mention their kids regardless of topic) and topics can vary widely (e.g., from public policy to personal leisure activities), so not all content words are useful across trials.  

Looking again at the heatmap and plots, we see that punctuation and double parentheses annotations are no longer relevant for the LDC transcription as they were in the ‘base’ setting. However, more character n-grams are important, such as “um”, “comp”, and “omp” (as in “computer”).

\begin{figure*}[!ht]
       \centering
       \includegraphics[width=.7\linewidth]{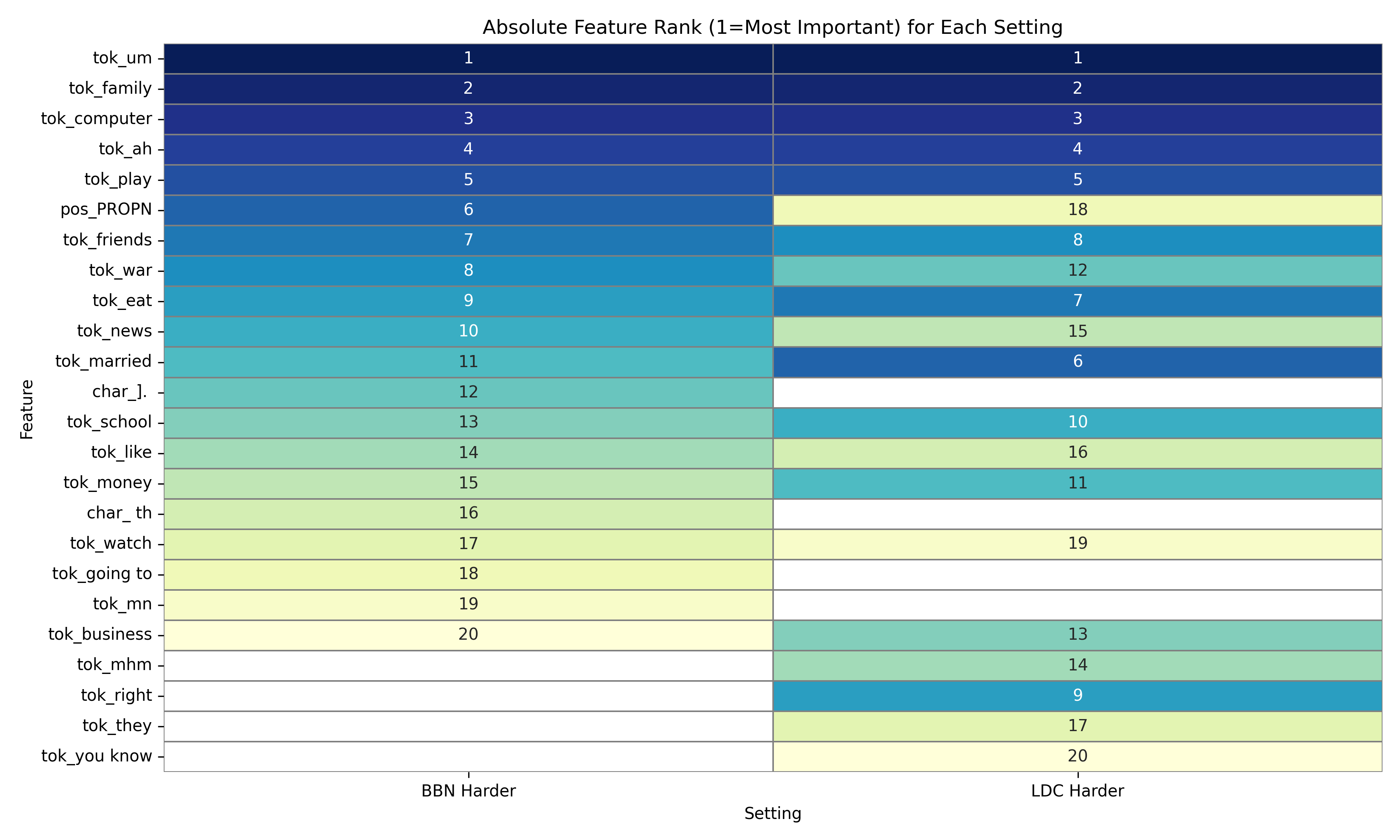}\\ \vspace{-.16cm}
      \includegraphics[width=0.49\linewidth]{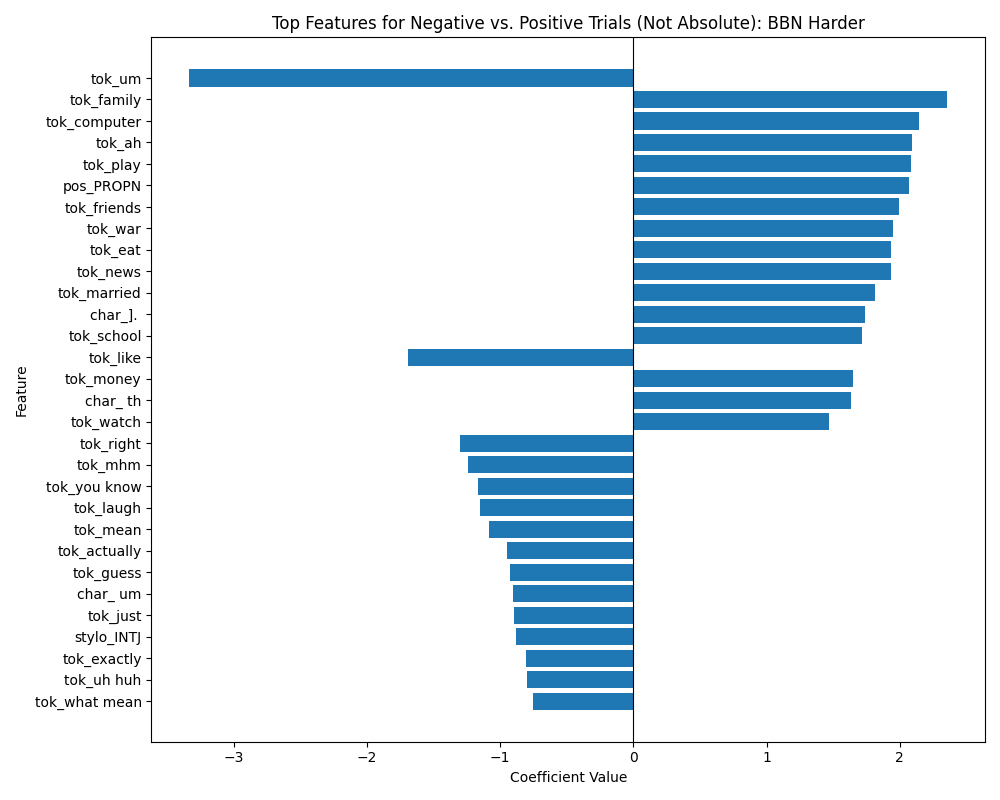}
      \includegraphics[width=0.49\linewidth]{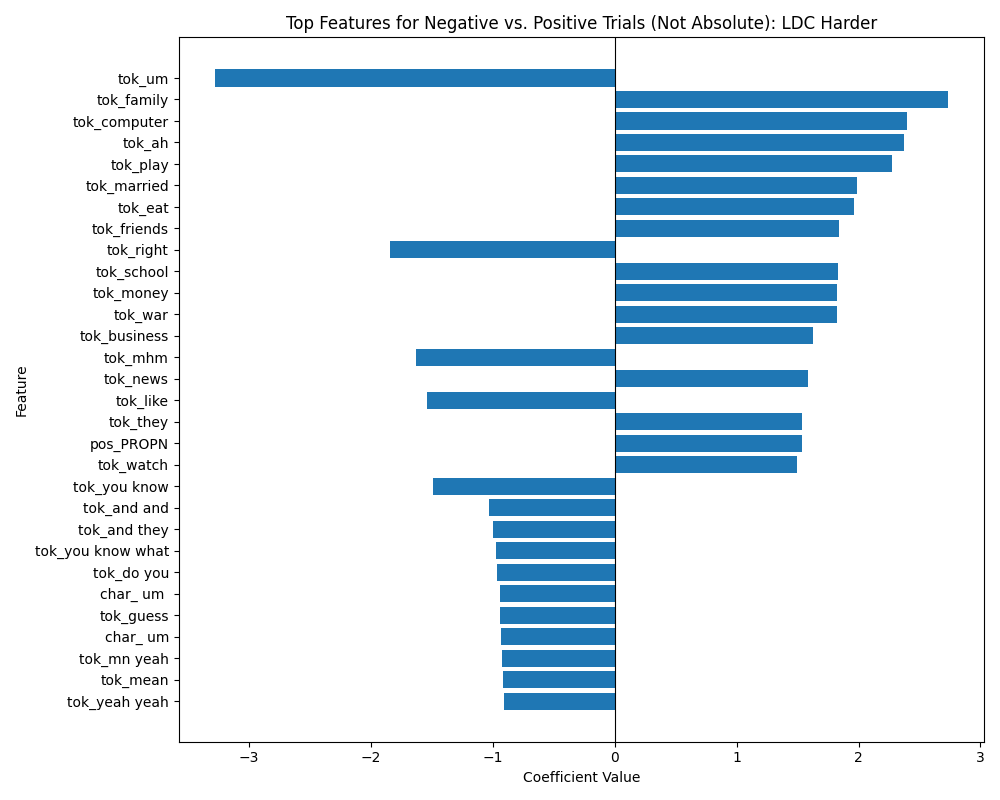}\vspace{-.2cm}
       \caption{\textbf{\emph{Top}}: Heatmap for the BBN (left) and LDC (right) transcription in the ‘\textbf{harder}’ difficulty level showing the ranking of the 20 most important absolute (not based on the logistic regression coefficient’s sign) features for the speaker verification task. Due to the stricter topic control setting, over half of the top features are content words for both transcriptions. \textbf{\emph{Bottom}}: Absolute ranking of the top 15 features for distinguishing negative (coefficients with a negative sign) and positive (coefficients with a positive sign) trials for BBN (left) and LDC (right). Overall, more features for distinguishing positive trials have larger absolute coefficients and are thus more predictive. 
        }
        \label{fig:harder}
   \end{figure*}

In the ‘harder’ level in \autoref{fig:harder}, BBN and LDC align on their top five top features, a mix of function words (\emph{um}, \emph{ah}) and content words (\emph{family}, \emph{computer}, \emph{play}). Again, most of the top token n-grams are content, not function words, and nearly all of the content words are useful for positive trails. More of these content words have larger absolute rankings in this difficulty level most likely because their differences are better predictors of same speaker trials in comparison to the different speaker trials, which involve speakers in the same call discussing the same topic and subtopics, thus having even more content word overlap. These starker contrasts likely contribute to this difficulty level having the highest performance. Proper nouns (PROPN) are now very relevant for BBN but less so for LDC. Looking at the POS tags for each transcription, there are significantly more proper nouns in BBN ($47,099$ in the test set) than in LDC ($21,793$). Since BBN includes capitalization and proper nouns are prescriptively capitalized, significantly more are captured, especially those that might be a common noun without context. For instance, both \emph{Fear} and \emph{Factor} are tagged as proper nouns in BBN when they are used to describe the television show Fear Factor, but tagged as nouns in LDC. BBN also now has a character n-gram, “].”, which is from brackets around non-speech sounds.

Overall, these feature importance results indicate a sensitivity to the topic control manipulations, especially in the ‘harder’ difficulty setting, but they otherwise align with previous studies that found function words to be important. More specifically, our results parallel those of Sari et al. \cite{sari2018}, who found that content words are better for high topical diversity data (such as our positive topic-controlled trials) and style features, including function words, are better for less topical diversity data (our negative topic controlled trials). These results diverge from studies that have found character n-grams to be important, which is reinforced by the results in \autoref{tab:all} showing that \textsc{PANgrams}, which specifically uses character 4-grams, did not perform nearly as well as \textsc{StyloSpeaker}. 

\section{Conclusion}\label{sec:concl}

Stylometric models based on textual features can be applied to speech transcripts productively, even if the transcript text has been normalized to remove text-like features, such as punctuation and capitalization. Taking the absolute difference of features for each trial provides the best speaker verification performance for our data and experimental setup. Performance is highest on the most topic-controlled setting, which involves different speaker trials of two speakers in the same conversation, most likely because the stylometric model learns to use both stylistic and content-based features to distinguish different speaker and same speaker trials, respectively. However, the neural models perform even better in this setting, as they are able to further use the topic information to their advantage. In the less topic-controlled settings, though, the stylometric model performs best, reaching AUC scores up to $0.86$. The most important features on this dataset in these topic-controlled settings were token n-grams, especially token unigrams. Across topic-control settings and transcription types, we consistently see some speech-related tokens among the most important, suggesting the possibility that laughter and discourse markers of various kinds may be under-utilized features in distinguishing the speech of different individuals. 

Future work could consider additional more complex features, such as syntactic dependency-based n-grams \cite{weerasinghe2020}, or different speech transcript datasets, including forensic ones, to see if the same features are relevant and if any feature paradigms exist for speech as a modality or specific speech registers. Although the higher performance of the explainable stylometric model over the black-box neural models in the ‘base’ and ‘hard’ settings is promising, we caution that this model should be applied carefully, especially in high-stakes forensic cases, with consideration for the amount of topic control and topic drift in the data.

\section{Declaration of generative AI use}
ChatGPT was used for troubleshooting the code development of the stylometric tool, but all suggestions were reviewed for accuracy. No generative AI tools were used for preparing the manuscript.

\section{Acknowledgments}
We thank Nicholas Andrews for insightful guidance and discussions on the design and analysis of the experiments.

\appendix
\section{Example of stylometric features}\label{app:feattable}
\autoref{tab:stylometry-features} shows the stylometric features and their frequencies for the example sentence ``\emph{Leave \$50,000 in the dumpster.}''.

\begin{sidewaystable}
\raggedright
\renewcommand{\arraystretch}{1.05} 
\resizebox{1.05\textwidth}{!}{
{\tiny
\begin{tabular}{|l|p{3cm}|p{8cm}|c|}
\hline
\textbf{Level} & \textbf{Stylometric Feature} & \textbf{"Leave \$\hspace{-.2em}50,000 in the dumpster."} & \textbf{Freq.} \\ \hline

\multirow{4}{*}{Character} 
& punctuation marks 
& ``,'', ``.'' (no others appear so have freq. of 0)
& 1,1,0,0,... \\ \cline{2-4}

& TF-IDF character n-grams (for n = 3, 4, 5, 6)
& \begin{tabular}[t]{@{}l@{}} 
3: ``lea'', ``eav'', ``ave'', ``ve '', ``e \$'', `` \$5'', ``\$50'', ``50,'', ``0,0'', ``,00'', ``000'', ``00 '',``0 i'',`` in'', ``in '', ``n t'', `` th'',... \\  
4: ``leav'', ``eave'', ``ave\ '', ``ve\ \$'', ``e\ \$50'', `` \$50'', ``\$50,'', ``50,0'', ``0,00'', ``,000'', ``000\ '', ``00\ i'', ``0\ in'', `` in '',... \\ 
5: ``leave'', ``eave\ '', ``ave\ \$'', ``ve\ \$5'', ``e\ \$50'', `` \$50,'', ``\$50,0'', ``50,00'', ``0,000'', ``,000\ '', ``000\ i'', ``00\ in'', ``0\ in\ '',... \\
6: ``leave\ '', ``eave\ \$'', ``ave\ \$5'', ``ve\ \$50'', ``e\ \$50,'', ``\ \$50,0'', ``\$50,00'',``50,000'', ``0,000 '', ``,000\ i'', ``000\ in'',... \\
\end{tabular}
& \\ \hline

\multirow{4}{*}{Token}
& \# of tokens (T) 
& ``leave'', ``\$'', ``50{,}000'', ``in'', ``the'', ``dumpster'', ``.'' 
& 7 \\ \cline{2-4}

& \# of unique tokens (U)
& ``leave'', ``\$'', ``50{,}000'', ``in'', ``the'', ``dumpster'', ``.'' 
& 7 \\ \cline{2-4}

& types:tokens (U:T)
& 7:7
& 1 \\ \cline{2-4}

& TF-IDF token n-grams (for n=1,2,3)
& \begin{tabular}[t]{@{}l@{}}
1: ``leave'', ``\$'', ``50{,}000'', ``in'', ``the'', ``dumpster'', ``.'' \\
2: ``leave\ \$'', ``\$ 50{,}000'', ``50{,}000 in'', ``in the'', ``the dumpster'', ``dumpster .'' \\
3: ``leave \$ 50{,}000'', ``\$ 50{,}000 in'', ``50{,}000 in the'', ``in the dumpster'', ``the dumpster .''
\end{tabular}
& \\ \hline

\multirow{4}{*}{Word}
& avg word length (\#chars)
& $5 + 2 + 3 + 8 = 18 / 4 = 4.5$
& 4.5 \\ \cline{2-4}

& short:W (<5 chars)
& $2:4 = 0.5$
& 0.5 \\ \cline{2-4}

& long:W ($\ge 8$ chars)
& $1:4 = 0.25$
& 0.25 \\ \cline{2-4}

& caps:W
& $1:4 = 0.25$
& 0.25 \\ \hline

\multirow{5}{*}{Syntax}
& \# of sentences
& ``Leave \$50{,}000 in the dumpster.'' 
& 1 \\ \cline{2-4}

& avg sent. length (\# toks)
& 7 / 1 = 7
& 7 \\ \cline{2-4}

& function word freq.
& in, the (no others appear so have freq. of 0)
& 1,1,0,0,... \\ \cline{2-4}

& function phrase freq.
& (none appear so all phrases have freq. of 0)
& 0,0,0,0... \\ \cline{2-4}

& POS tag freq. 
& VERB, SYM, NUM, ADP, DET, NOUN, PUNCT 
& 1,1,1,1,1,1,1 \\ \cline{2-4}

& TF-IDF POS n-grams (for n=1,2,3)
& \begin{tabular}[t]{@{}l@{}}
1: VERB, SYM, NUM, ADP, DET, NOUN, PUNCT \\
2: (VERB SYM), (SYM NUM), (NUM ADP), (ADP DET), (DET NOUN), (NOUN PUNCT) \\
3: (VERB SYM NUM), (SYM NUM ADP), (NUM ADP DET), (ADP DET NOUN), (DET NOUN PUNCT)
\end{tabular}
& \\ \hline

\multirow{6}{*}{Discourse}
& vocab. richness (Yule's I)
& $(1+1+1+1+1)^2 / ((1^2+1^2+1^2+1^2+1^2)) = 25/0 \rightarrow \text{ZeroDivision} : I = 0$
& 0 \\ \cline{2-4}

& readability measures
& \begin{tabular}[t]{@{}l@{}}
Flesch Reading Ease \\
SMOG Index \\
Flesch–Kincaid Grade \\
Coleman–Liau Index \\
Automated Readability Index \\
Dale–Chall Score \\
Difficult Words \\
Linsear Write Formula \\
Gunning Fog Index
\end{tabular}
& 
\begin{tabular}[t]{r}
100.24 \\
3.1291 \\
0.52 \\
4.96 \\
5.56 \\
10.20 \\
1 \\
1.5 \\
2.0
\end{tabular}\\ \cline{2-4}

& hapax legomena:W
& $5:5 = 1$
& 1 \\ \cline{2-4}

& hapax dislegomena:W
& $0:5 = 0$
& 0 \\ \cline{2-4}

& \# of contracted terms
& 0
& 0 \\ \cline{2-4}

& \# non-contracted terms
& 0
& 0 \\ \hline

\end{tabular}}}
\caption{Example stylometric feature extraction results for the utterance ``Leave \$50{,}000 in the dumpster.''. The character n-grams include spaces. The TF-IDF n-grams do not have frequencies in the table because their values are calculated in relation to how often they appear in a speaker's utterances and the corpus overall. Hapax legomena/dislegomena is calculated within a speaker's utterances in a particular call, not in relation to the corpus overall. Here, we consider the utterance to be the whole call for demonstration purposes.}
\label{tab:stylometry-features}
\end{sidewaystable}

\section{Accuracy and EER}\label{app:acceer}
\autoref{tab:acceer} shows the corresponding results across feature measurements and difficulty levels using the metrics accuracy (top) and equal error rate (EER; bottom). Accuracy (higher values are better) measures the number of correct predictions out of all predictions the classifier made and works best for classification when the positive and negative classes are balanced, which is the case for this dataset. EER (lower values are better) is the point when the false acceptance rate equals the false rejection rate, or when the probability of incorrectly classifying a negative trial as positive equals the probability of incorrectly classifying a positive trial as negative. Unlike accuracy, which only measures whether predictions are correct or not, EER takes into account which class is being misclassified, making it potentially more informative in evaluating performance. 

In both tables, the best overall feature measurement (among \emph{Concat}, \emph{Diff}, \emph{Diff + Prod}, and \emph{Concat + Diff}) is taking the absolute difference and product (\emph{Diff + Prod}) of the features on each side of a trial. This differs from AUC, where \emph{Diff} alone was the top-performing measurement. \emph{Diff} captures how much the two sides of a trial differ, while \emph{Prod} reveals how much they align. Using both together improves accuracy and EER because these metrics evaluate performance at a specific threshold (generally $0.5$ for accuracy and where false acceptance and false rejection are equal for EER). The product term adds extra information that helps separate the classes more clearly. By contrast, AUC measures how well the model ranks positive trials higher than negative ones across all possible thresholds, so the product term might not contribute additional discerning information in that context.   

\begin{table}[ht]
\centering
\begin{tabular}{|l|cccc|cccc|}
\hline
\multirow{2}{*}{\textbf{Acc $\uparrow$}} 
  & \multicolumn{4}{c|}{\textbf{BBN}} 
  & \multicolumn{4}{c|}{\textbf{LDC}} \\
\cline{2-9}
  & \textit{Concat} & \textit{Diff} & \textit{Diff+Prod} & \textit{Concat+Diff}
  & \textit{Concat} & \textit{Diff} & \textit{Diff+Prod} & \textit{Concat+Diff} \\
\hline
\textbf{Base}   & 0.528 & 0.535 & \textbf{0.697} & 0.687 
                & 0.520 & 0.536 & \underline{0.696} & 0.694 \\
\textbf{Hard}   & 0.543 & 0.523 & 0.643 & 0.650 
                & 0.542 & 0.529 & \textbf{0.681} & \underline{0.669} \\
\textbf{Harder} & 0.494 & 0.505 & \textbf{0.746} & \underline{0.734} 
                & 0.498 & 0.498 & 0.728 & 0.696 \\
\hline
\end{tabular}

\vspace{2ex}

\begin{tabular}{|l|cccc|cccc|}
\hline
\multirow{2}{*}{\textbf{EER $\downarrow$}} 
  & \multicolumn{4}{c|}{\textbf{BBN}} 
  & \multicolumn{4}{c|}{\textbf{LDC}} \\
\cline{2-9}
  & \textit{Concat} & \textit{Diff} & \textit{Diff+Prod} & \textit{Concat+Diff}
  & \textit{Concat} & \textit{Diff} & \textit{Diff+Prod} & \textit{Concat+Diff} \\
\hline
\textbf{Base}   & 0.470 & 0.469 & \underline{0.302} & 0.314 
                & 0.475 & 0.467 & \textbf{0.301} & 0.308 \\
\textbf{Hard}   & 0.463 & 0.479 & 0.355 & 0.359 
                & 0.467 & 0.479 & \textbf{0.321} & \underline{0.330} \\
\textbf{Harder} & 0.498 & 0.493 & \textbf{0.258} & \underline{0.261} 
                & 0.491 & 0.491 & 0.266 & 0.299 \\
\hline
\end{tabular}
\caption{Accuracy (top) and equal error rate (bottom) across feature combination methods for BBN and LDC. Best values are bolded; second-best are underlined. Taking the absolute difference and product (\emph{Diff + Prod}) of the features is the best overall measurement for both metrics.}
\label{tab:acceer}
\end{table}

\bibliographystyle{model1-num-names}

\bibliography{custom}

\end{document}